\documentclass{article}
\usepackage{spconf,amsmath,graphicx}
\usepackage[english]{babel}
\usepackage{multirow}

\title{Pronunciation-aware unique character encoding for RNN Transducer-based Mandarin speech recognition}
\name{Peng Shen, Xugang Lu, Hisashi Kawai}
\address{National Institute of Information and Communications Technology (NICT)}
\begin{document}
%
\maketitle
\begin{abstract}
For Mandarin end-to-end (E2E) automatic speech recognition (ASR) tasks, compared to character-based modeling units, pronunciation-based modeling units could improve the sharing of modeling units in model training but meet homophone problems.
In this study, we propose to use a novel pronunciation-aware unique character encoding for building E2E RNN-T-based Mandarin ASR systems. The proposed encoding is a combination of pronunciation-base syllable and character index (CI). By introducing the CI, the RNN-T model can overcome the homophone problem while utilizing the pronunciation information for extracting modeling units. With the proposed encoding, the model outputs can be converted into the final recognition result through a one-to-one mapping.
We conducted experiments on Aishell and MagicData datasets, and the experimental results showed the effectiveness of the proposed method.

\end{abstract}
\section{Introduction}
\label{sec:intro}
Recently, end-to-end (E2E) automatic speech recognition (ASR) techniques have achieved significant progress with the development of system architecture and optimization algorithms \cite{jinyuLi2021E2EASR}.
Previous works have shown that the performance of the recently advanced E2E system competes with or even outperforms that of traditional ASR approaches \cite{Sainath2020ASO, Li2020DevelopingRM}.
Among the E2E techniques \cite{Li2020DevelopingRM, Chorowski2014EndtoendCS, Chan2016ListenAA, Alex2012RNNT}, recurrent neural network Transducer (RNN-T) \cite{Alex2012RNNT} is one of the most popular E2E models for its natural streaming capability and performance.
This work also focuses on using RNN-T models for Mandarin speech recognition tasks.

In traditional hybrid ASR systems, context-dependent phonemes (CD-phone) are widely used as the modeling units. A lexicon-based converter,  for example, the weighted finite state transducer (WFST) decoder, is required to convert the predicted outputs to the text sequences \cite{Yu2014ASR}.
For E2E models, the predicted outputs are expected to be the final recognition results. Therefore, comparing CD-phone and phone, characters, words, and sub-words are more widely used as modeling units \cite{Graves2014TowardsES, Hannun2014DeepSS, Sennrich2016NeuralMT, Audhkhasi2018BuildingCD, Zenkel2018SubwordAC}.
Recently, most of the state-of-the-art ASR system on English use sub-word techniques, for example sentencepiece \cite{Sennrich2016NeuralMT}, to extract the modeling units \cite{conformer2020, Zhao2021OnAP}.

In Mandarin E2E ASR models, various modeling units, such as character, word, syllable, and CD-phone, have also been explored \cite{Zhang2019InvestigationOM, Chen2021AnIO, Zou2018ComparableSO, Wang2021CascadeRS}.
Previous works show that models using a combination of characters and syllables outperform those using only characters \cite{Zhang2019InvestigationOM, Zou2018ComparableSO}.
However, because of the homophone problem in Mandarin, we need a further process to convert the predicted outputs to final transcriptions when using syllables and phonemes as units \cite{Wang2021CascadeRS}. This makes the E2E model lose the end-to-end characteristics.
Therefore, how to use pronunciation-aware modeling units for building E2E Mandarin ASR is still an open topic.

Considering the pronunciation information underlying acoustic signal is an essential cue for ASR tasks, therefore, taking the pronunciation information into account when designing modeling units is necessary.
In this work, we propose to design a pronunciation-aware unique character encoding method for building E2E Mandarin ASR systems.
The proposed encoding consists of three partitions: syllable, tone, and character index (CI).
The tonal syllable can be optimized with the acoustic feature, and the character index can be optimized and enhanced with text corpus-based linguistic features.
Unlike previous works that need a convertor when using pronunciation-based modeling units, models based on the proposed encoding can run entirely in E2E mode.
The contributions of this work can be summarized as follows:
\begin{itemize}
  \item We analyze the characteristics of the RNN-T and propose a novel pronunciation-aware unique character encoding method for RNN-T-based Mandarin ASR tasks.
  \item To reduce the prediction difficulty of RNN-T, we propose to use meta symbols rather than integers for tone and CI. The experimental results show that the particular meta symbol-based encoding makes the ASR model perform better.
  \item We further investigate a CI-based beam searching for external LM rescoring and show the potential of the proposed encoding in solving the low computational efficiency problem of beam searching and the difficulty of exploiting text corpus in the RNN-T approach.
\end{itemize}
We evaluated the proposed method on 165-hour Aishell and 755-hour MagicData Mandarin datasets by comparing other fully E2E modeling units. The experimental results showed that the proposed method outperformed the character- and char-word-based baseline systems with an E2E mode.

\section{RNN Transducer}
\begin{figure}
  \centering
  \includegraphics[width=250px]{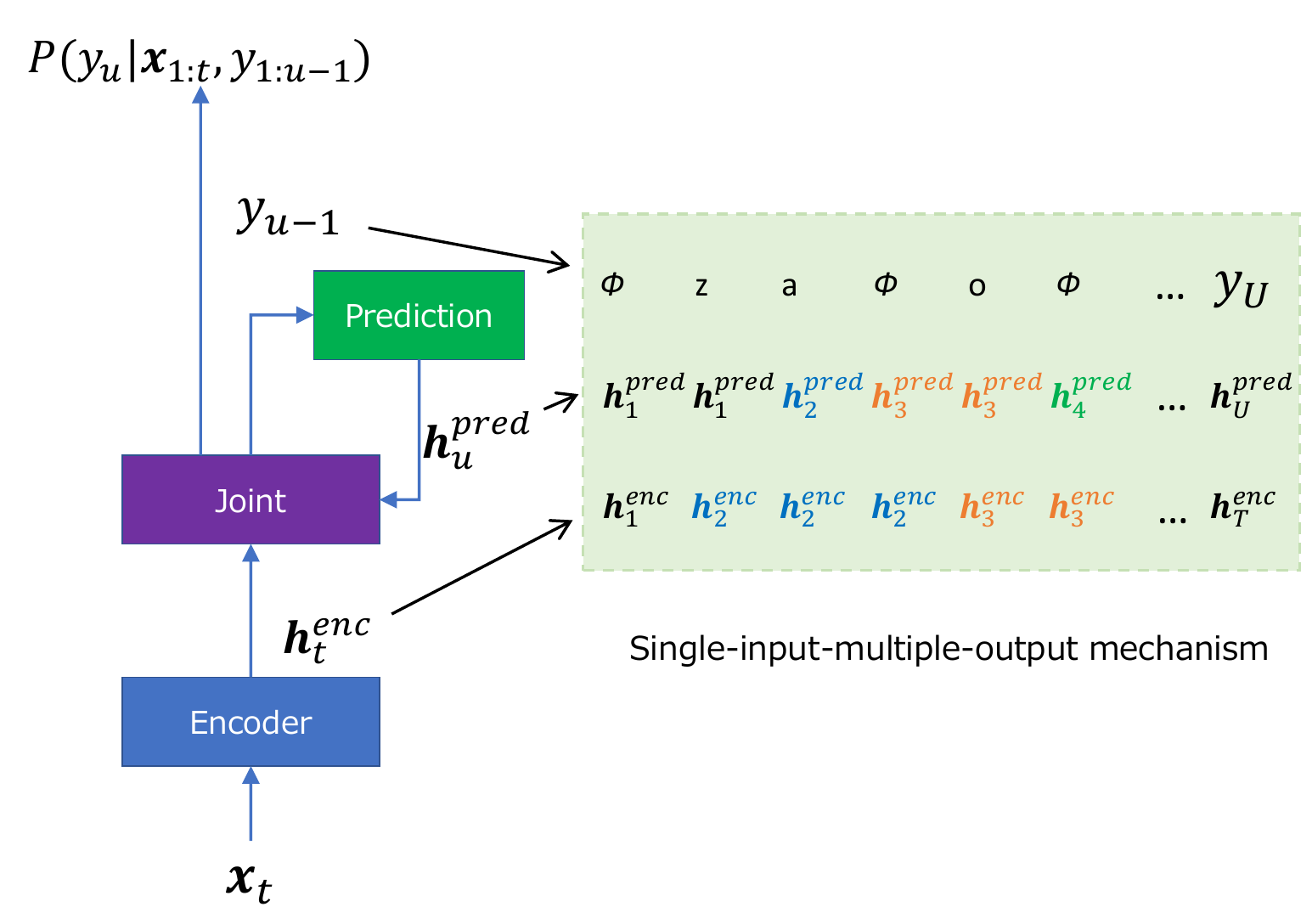}\\
  \caption{RNN Transducer and its single input multiple output mechanism.}\label{fig.rnnt}
\end{figure}

Compared with the conventional connectionist temporal classification (CTC), RNN-T removes the conditional independence assumption, and it outputs conditions on the previous output tokens and the speech sequence until the current time step as $P(y_u|\mathbf{x}_{1:t},y_{1:u-1})$.
As illustrated in Fig. \ref{fig.rnnt}, an RNN-T consists of an encoder, prediction, and joint network.
The encoder network plays an acoustic feature extraction role by generating a high-level feature representation $\mathbf{h}^{enc}_t$.
The prediction network is a language model that produces a linguistic representation $\mathbf{h}^{pre}_u$ based on RNN-T's previous output label $y_{u-1}$.
Then $\mathbf{h}^{enc}_t$ and $\mathbf{h}^{pre}_u$ are combined with the feed-forward network-based joint network to predict the output symbols.

Because the output sequence length is usually shorter than the input in ASR tasks, an RNN-T model allows the output of a special blank symbol $\phi$, which skips this input frame without changing the state of the prediction network.
That means the blank symbol decides whether to move to the next time frame $t+1$ or to emit more output units from the same time frame $t$ for the next joint network call.
As illustrated in Fig. \ref{fig.rnnt}, the RNN-T is a single-input-multiple-output device that outputs one or more symbols (including the blank sign) at any given time $t$ given the audio features up to $x_t$.

\section{RNN-T with unique character encoding}
\label{sec:page}
Unlike the CTC-based models, which predict only one output symbol for every acoustic input, the RNN-T model is a single-input-multiple-output model. This characteristic motivates us to consider whether the RNN-T model can further output information more than symbols related to pronunciation information.
In this work, we assume that the RNN-T model can further predict more information after the output related to the acoustic feature. Under this assumption, we define a pronunciation-aware unique character encoding (PUCE) for building RNN-T-based Mandarin ASR systems so that our RNN-T model can run on an E2E mode.


\subsection{Definition}
\label{sec:definition}
\begin{figure}
  \centering
  \includegraphics[width=230px]{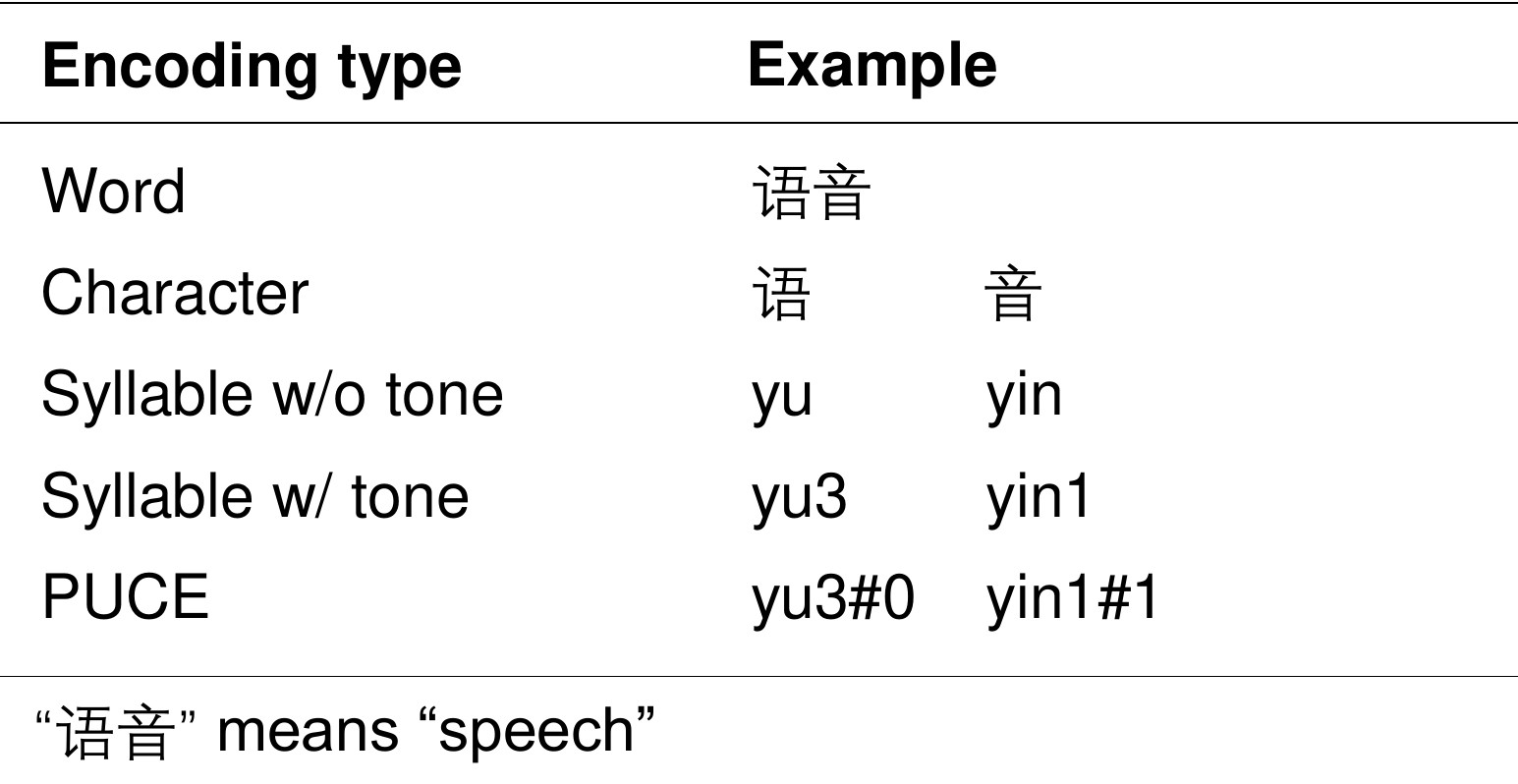}\\
  \caption{Examples of different modeling unit types.}\label{fig.units.example}
\end{figure}

Fig. \ref{fig.units.example} illustrates a modeling unit example for a Chinese word that means "speech." To build a fully E2E system, we can select characters or words as modeling units.
However, because of the homophone problem, pronunciation-based units can not fully support E2E processing mode because one tonal syllable corresponds to one to hundreds of characters in Chinese.
Motivated by previous works on sub-character tokenization on NLP tasks \cite{Si2021SubCharacterTF, Zhang2021UsingSL}, we define a syllable-based unique character encoding method to represent the character for ASR tasks.
For a Chinese character $C$, let $p$ be the character's pronunciation, for example, a syllable, and $TI$ is its tone information since Mandarin is a tonal language. Let $CI$ be the $I$-th character of the corresponding tonal pronunciation $(p, TI)$. Then, the encoding can be described as
\begin{equation}\label{Definication of UCE}
  E_{C} = ((p, TI), CI).
\end{equation}
In this definition, $p$ can use phonemes, syllables, or IPA-inspired phonemes.
Mandarin has five tones: first, second, third, fourth, and neutral; therefore, we define the $TI \in \{1,2,3,4,5\}$ that corresponds to the five tones.
We can remove the $TI$ for languages without tone information.

To reduce the represent confusion of $TI$ and $CI$, rather than using integers, we use meta symbols for $TI$ and $CI$ by referring to the space in sentencepiece.
We selected Unicode code points from 10,049 to 10,054 (five tones) for $TI$ and 41,000 to 41,900 (support maximum character index of 900) for $CI$.


\subsection{E2E RNN-T pipeline}
\begin{figure}
  \centering
  \includegraphics[width=230px]{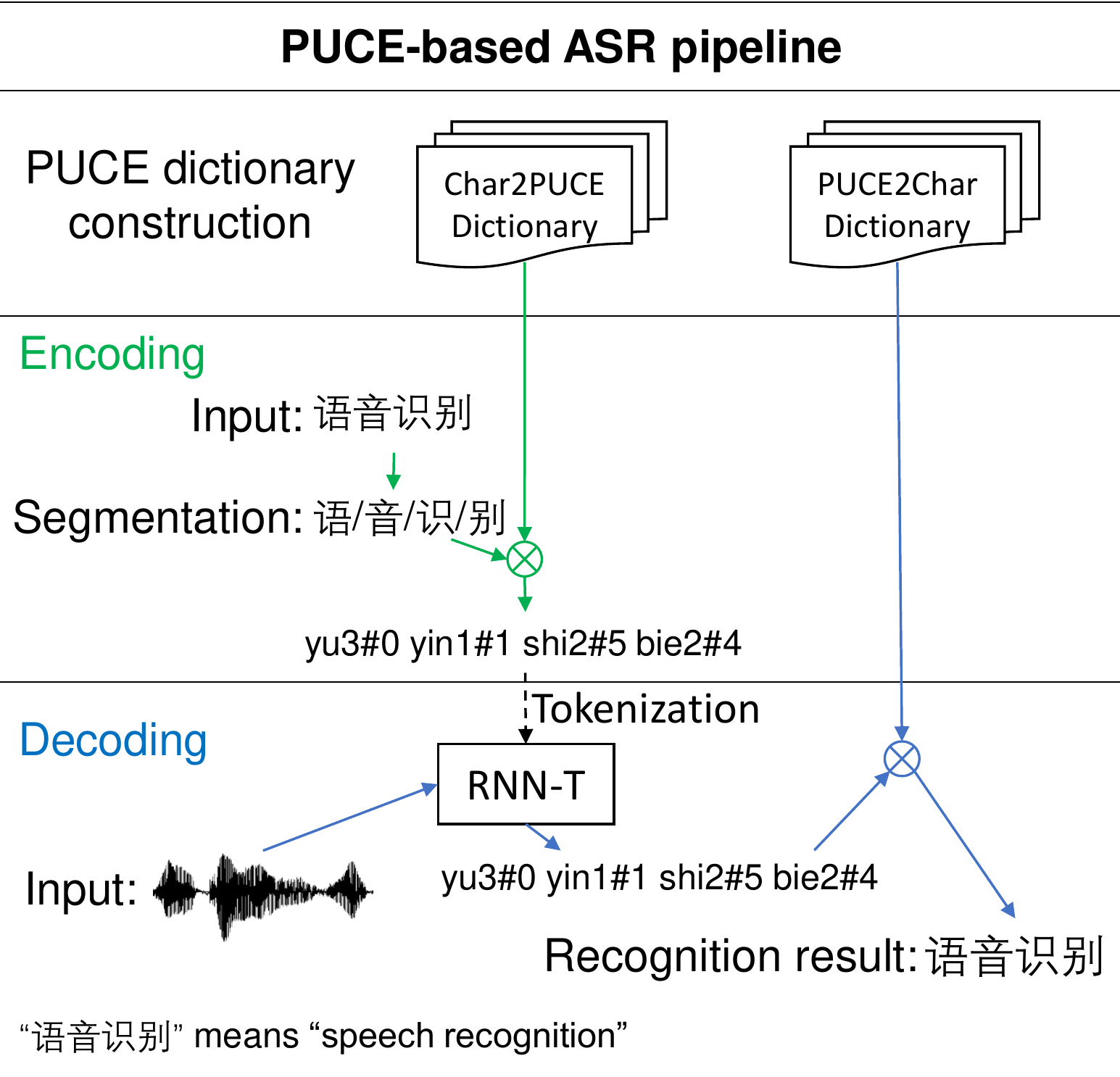}\\
  \caption{The proposed PUCE-based ASR pipeline.}\label{fig.pipeline}
\end{figure}

Fig. \ref{fig.pipeline} illustrates the whole pipeline of RNN-T-based ASR using the proposed encoding method.
Firstly, we need to prepare two dictionaries: an encoding dictionary for encoding characters to symbol lists and another for changing the predicted symbols to characters.
For Mandarin, not only does a pronunciation correspond to multiple characters, but also some characters have more than one pronunciation. Therefore, the construction of the two dictionaries is described as follows:
Given a tonal syllable $S_1$ and its corresponding characters $C_1, C_2, \cdots, C_i$, the decoding dictionary is created with $\{S_1\#1:C_1, S_1\#2: C_2, \cdots , S_1\#i: C_i \}$. Then the encoding dictionary is created by changing the place of key and value of the decoding dictionary to $\{C_1:S_1\#1, C_2:S_1\#2, \cdots , C_i: S_1\#i\}$.
For a character $C_3$ with multiple tonal syllables $S_1, S_2$, its value is represented as $\{C_3:S_1\#1; S_2\#2 \}$ in the encoding dictionary.
With the proposed decoding dictionary, we change the one-to-many mapping problem to one-to-one mapping in the inference stage to overcome the homophone problem.

With the prepared encoding dictionary, the text of training data is encoded by a simple lookup operation following an operation to split the input sentences into characters.
After that, we use tokenization techniques to extract the modeling units for the RNN-T model training.
In the inference stage, similar to the process for English, the symbols of RNN-T outputs are first processed with the tokenization model to obtain the proposed encoding.
Finally, a lookup operation is done to obtain the final recognition results with the decoding dictionary.

\subsection{Modeling units with tokenization}
Word piece/sub-word tokenization techniques have been widely used in E2E ASR tasks.
Different from the character- and char-word-based methods, the proposed encoding can use tokenization to extract sub-char-based tokens as modeling units. In this work, we apply tokenization on the pronunciation partition, i.e., $(p, TI)$.
With tokenization, the higher frequency syllable is represented as a unique symbol, and the less frequent syllable is split into sub-tokens.
This helps the model not to learn less frequent words and is expected to identify the words that do not exist in the training data set.

\subsection{CI-based beam searching and rescoring}
\label{sec:fusion}
The performance of the ASR model can be further improved by using external LMs trained on text-only data.
In traditional hybrid ASR systems, rescoring with a neural network-based language model (NN-LM) on the N-best hypothesis lists is widely used.
Because the RNN-T is an autoregressive model, the computational efficiency for extracting the N-best hypothesis is poor, especially with a larger beam width. Therefore, the shallow fusion technique during the inference stage is more widely used \cite{Cabrera2021LanguageMF, Kim2021ImprovedNL}.

In the proposed method, benefitting from the separated definition of pronunciation and the CI, we can implement CI-based beam-searching for extracting N-best lists.
In detail, we use the top-1 syllable and tone outputs and execute beam searching on the CI outputs.
With the modified N-best list, external LM is used to reorder the hypothesis as
\begin{equation}\label{eq.fusion}
  \hat{y} = \log P_{RNNT}(y|x) + \lambda \log P_{LM}(y),
\end{equation}
where $P_{RNNT}(y|x)$ is the posterior output probability of RNN-T, and $P_{LM}(y)$ is the probability of $y$ generated by external LMs.
Because we use unique meta symbols for $CI$ rather than the integer, there is no autoregressive process during the beam searching, and no penalty term on the tokens count since the number of tokens is identical for all the N lists.

\section{Experiments and results}
\subsection{Dataset}
Experiments were conducted on the Aishell-1 \cite{aishell2017} and MagicData \cite{magicdata2019} Mandarin speech corpus.
The Aishell dataset contains about 150 hours of speech recorded by 340 speakers for the training set. The development set includes about 10 hours recorded by 40 speakers. And about 5 hours of speech were recorded by 20 speakers as the test set.
We applied speed perturbation in the time domain with factors 0.9 and 1.1 to augment the training data to around 450 hours for model building.
The MagicData contains 755 hours of scripted read speech data from 1080 native speakers of Mandarin Chinese spoken in mainland China.
The database was split into a training set, validation set, and testing set in a ratio of 51: 1: 2.

\subsection{Experimental setup}
Because our target was building a fully E2E ASR system, we built a character-based RNN-T model as a baseline.
The number of character modeling units was 4,232 for the Aishell dataset and 4,479 for the MagicData dataset.
As previous works showed the effectiveness of char-word units \cite{Chen2021AnIO}, we also built a char-word-unit-based baseline system using the sentencepiece toolkit \cite{Kudo2018SentencePieceAS} to prepare tokens with the token number set to 5,000 and 8,000.
The same tokenization method was used for the proposed approach to extract the modeling units by taking the $TI$ and $CI$ as special symbols.
Performances with the different token numbers, i.e., 191/208, 300, 500, and 1,000, were compared.

We used the conformer network for the RNN-T encoder and one LSTM layer for the prediction network by referencing to \cite{conformer2020}.
The conformer encoder consists of 16 conformer blocks with an encoder dimensional of 176, attention heads set to 8, and convolutional kernel number set to 32. Subsampling with factor 4 was processed before the conformer layers.
The prediction network consists of 320 LSTM memory cells.
The outputs of the encoder and prediction network were transformed to 320 dimension vectors with a linear transform layer of the joint network. The neurons of the final fully-collected layer of the joint network were 320.
The input acoustic feature was an 80-dimensional log Mel-based filter-bank feature extracted with 25 ms frame length and 10 ms frameshift.
SpecAugment technique was applied during the model training to improve the robustness of the model \cite{Park2019SpecAugment}.
All the models were trained from scratch. During model training, we used the NovoGrad algorithm \cite{Ginsburg2019StochasticGM} with a learning rate of $0.03$ and a cosine annealing warm restart learning rate schedular \cite{Loshchilov2016Warmup} to adjust the learning rate with the minimum learning rate was set to 1e-6.
The mini-batch size was 64. The number of training epochs was set to 100.

Two transformer LMs trained on Aishell and MagidData text were used for N-best list rescoring.
The LMs were trained with tokens obtained based on the proposed encoding.
The hidden dimension of the transformer LM is 512, except for the feedforward sublayer of size 2048. We set the number of heads to 8 and stack 6 layers in the encoder. The dropout was set to 0.1.
The Adam optimizer with a warmup setting with a learning rate of 0.001 was used.
The training epochs were set to 200.
All the methods were implemented with the NVIDIA NeMo\cite{kuchaiev2019nemo} toolkits.
The experimental results were reported in character error rate (CER \%).

\subsection{Investigation and experimental results}

\subsubsection{Modeling units with characters and words}
For building E2E Mandarin ASR, using CD-phone and syllable as modeling units often need a complex converter to convert predicted symbols to characters.
In this work, we investigated modeling units with characters and a combination of characters and words for fully E2E ASR baseline systems.
Table \ref{result:baseline} shows the experimental results for these baselines on both Aishell and MagicData datasets.
We also listed some results reported by other researchers for reference.
From the results, we can see that compared to using characters, the combination of characters and high-frequency words showed better performance. However, with the increase in the number of words selected, the performance was degraded.

\begin{table}[tb]
\centering
\caption{Results (CER \%) of the baseline systems on Aishell and MagicData.}
\setlength{\tabcolsep}{0.4em}
\begin{tabular}{|l|c|c|c|c|c|c|} \hline
Dataset-ModelingUnits                & Dev  & Test   \\ \hline \hline
Aishell char(4232)          & 6.09  & 6.49  \\ \hline
Aishell char-word(5000)     & \textbf{5.90}  & \textbf{6.48}  \\ \hline
Aishell char-word(8000)     & 6.30  & 6.98  \\ \hline
Aishell syllable-char-word \cite{Chen2021AnIO}          & 6.02 & 6.73 \\ \hline
Aishell char \cite{Luo2021SimplifiedSF}     & - & 6.84 \\ \hline\hline

MagicData char(4479)            & 5.79 & 5.78 \\ \hline
MagicData char-word(5000)       & \textbf{4.97} & \textbf{3.76} \\ \hline
MagicData char-word(8000)       & 5.62 & 4.22 \\ \hline
MagicData syllable-char-word \cite{Chen2021AnIO} & 5.38  & 5.51  \\ \hline

\end{tabular}
\vspace{-3mm}
\label{result:baseline}
\end{table}

\subsubsection{Investigations on $TI$ and $CI$}
The $TI$ and $CI$ of the proposed encoding can use integers. However, because the $TI$ is related to pronunciation information and $CI$ is associated with the character index, therefore, using the same integer to represent may improve confusion between them; for example, the $TI=1$ means first tone, but $CI=1$  means the second character for a tonal syllable.
We compared integers and meta symbols (MS) (as described in section \ref{sec:definition}) for $TI$ and $CI$ to investigate the influence.
The investigation results are listed in Table \ref{result.investigate1}.
From the results, we can see that using integers for both $TI$ and $CI$, the RNN-T model worked well even though the number meaning was different.
When using integer symbols for $CI$, performance on both datasets degraded as the number of tokens increased.
When using MS for $CI$, with the increasing of the token number, the performance could be improved on the relatively large dataset, i.e., MagicData, and obtained almost the same error rate on the Aishell dataset.

\begin{table}[tb]
\centering
\caption{Investigation of symbols for $TI$ and $CI$ (INT means integer, MS means meta symbols, TN means token number, MIN means minimum token number).}
\setlength{\tabcolsep}{0.5em}
\begin{tabular}{|l|c|c|c|c|c|c|c|c|} \hline
Dataset    & $TI$ & $CI$ &$TN$            & Dev  & Test    \\ \hline \hline

Aishell &INT & INT & 38(MIN)         & 5.63   & \textbf{6.24}   \\ \hline
Aishell &MS  & INT & 43(MIN)        & 5.71   & 6.32   \\ \hline
Aishell &MS  & INT & 100       & 6.70   & 6.34   \\ \hline
Aishell &MS  & INT & 300       & 5.83   & 6.51   \\ \hline
Aishell &MS  & MS & 191(MIN)        & 5.63   & \textbf{6.24}   \\ \hline
Aishell &MS  & MS & 300        & \textbf{5.60}   & 6.25   \\ \hline
Aishell &MS  & MS & 500        & 5.76   & 6.31   \\ \hline
Aishell &MS  & MS & 1000        & 5.68   & 6.29   \\ \hline\hline

MagicData &INT& INT & 54(MIN)       & 5.31   & 4.06 \\ \hline
MagicData &MS & INT & 59(MIN)       & 4.95   & 3.51 \\ \hline
MagicData &MS & INT & 100      & 5.25   & 3.81 \\ \hline
MagicData &MS & INT & 300      & 5.65   & 4.48 \\ \hline
MagicData &MS  & MS & 208(MIN)      & 5.40   & 4.08   \\ \hline
MagicData &MS  & MS & 300      & 5.31   & 3.83   \\ \hline
MagicData &MS  & MS & 500      & 4.86   & 3.53   \\ \hline
MagicData &MS  & MS & 1000      & \textbf{4.74}   & \textbf{3.46}   \\ \hline

\end{tabular}
\vspace{-3mm}
\label{result.investigate1}
\end{table}

\subsubsection{CI-based beam searching and rescoring}
As described in section \ref{sec:fusion}, models with the proposed encoding can use CI-based beam searching, and then the extracted N-best can be reordered with an external language model.
Different from the beam searching of the vanilla RNN-T, the proposed CI-based beam searching does not need an autoregressive operation; therefore, the searching has the same computational complexity as the CTC-based beam searching.
We listed the rescoring results in Table \ref{result.summary}.
For Aishell test data, we used the transformer LM that trained on Aishell text data.
And rescoring on MagicData test data, the LM trained on MagicData text was used.
The $\lambda$ was determined based on the dev datasets, which were 1.16 and 2.08 for the in- and cross-domain dataset of Aishell-PUCE(191), and were 0.55 and 1.85, 0.61 and 1.50 for MagicData-PUCE(208) and MagicData-PUCE(1000), respectively.
From the results, we can see that rescoring the CI-based N-best could further improve the performance on both the in-domain and the cross-domain datasets.


\begin{table}[tb]
\centering
\caption{Summary of the baselines and the proposed method (PUCE) on in- and cross-domain datasets (CER \%).}
\setlength{\tabcolsep}{0.45em}
\begin{tabular}{|l|c|c|c|c|c|c|c|c|} \hline
\multirow{1}{*}{} & \multicolumn{2}{c|}{InDomain} & \multicolumn{2}{c|}{CrossDomain} \\ \cline{2-5}

Method(Training data - units)                  & Dev & Test  & Dev & Test   \\ \hline \hline

Aishell-char(4232)              & 6.09 & 6.49  & 40.29 &  30.98  \\ \hline
Aishell-char-word(5000)         & 5.90 & 6.48  & 35.39 &  30.43  \\ \hline
Aishell-PUCE(191)               & 5.63 & 6.24  & 33.72 &  27.66  \\ \hline
\quad \quad \quad + CI-based rescoring & \textbf{5.33} & \textbf{5.92}  & \textbf{30.88} &  \textbf{24.38}  \\ \hline\hline

MagicData-char(4232)            & 5.79 & 5.78  & 24.63  &  26.96  \\ \hline
MagicData-char-word(5000)       & 4.97 & 3.76  & 25.24  &  27.76  \\ \hline
MagicData-PUCE (208)            & 5.40 & 4.08  & 20.78  &  22.88  \\ \hline
\quad \quad \quad + CI-based rescoring & 5.26 & 3.93  & \textbf{17.25} &  \textbf{19.20}  \\ \hline
MagicData-PUCE (1000)           & 4.74 &3.46   & 21.97 &  24.05  \\ \hline
\quad \quad \quad + CI-based rescoring & \textbf{4.60} & \textbf{3.32} & 18.35  &  20.37    \\ \hline

\end{tabular}
\vspace{-3mm}
\label{result.summary}
\end{table}

\subsubsection{Performance on cross-domain datasets.}
In Table \ref{result.summary}, we listed the investigation on the cross-domain tasks.
We evaluated the MagicData test data with the model trained on the Aishell dataset and evaluated Aishell test data with the model trained on MagicData.
Using a combination of characters and words could improve the performance on the in-domain dataset; however, for the unseen domain, the performance even changed worse.
The proposed method outperformed the two baseline systems on both the in and cross-domain datasets.
Experimental results also show that as the number of tokens increases, the performance decreases on cross-domain dataset.


\subsection{Summary and discussion}

In this work, considering the autoregressive characteristics of RNN-T, we defined a pronunciation-aware unique character encoding method for building E2E Mandarin ASR systems.
To reduce the confusion and improve the efficiency of model optimization, we further enhanced the encoding by using meta symbols for both $TI$ and $CI$.
We evaluated the proposed method and showed the effectiveness of the proposed method for the Mandarin ASR tasks.
Further investigations on cross-domain datasets and CI-based N-best rescoring illustrated the robustness and the potential of exploiting the external text corpus of the proposed method.

In this work, we implemented CI-based N-best rescoring with the transformer LMs.
The popular shallow fusion \cite{Cabrera2021LanguageMF, Kim2021ImprovedNL} can also be used with the proposed modeling units. And CI-based rescoring based on the results with shallow fusion could further improve the performance.
As the char-word baseline system showed, taking high-frequency words as tokens could improve the performance on in-domain tasks. Therefore, further designing word-based encoding will be one of our future works.


\section{Conclusions}
In this work, we proposed to use a novel pronunciation-aware unique character encoding for building E2E RNN-T-based Mandarin ASR systems.
The proposed encoding was a combination of pronunciation-base syllable and character index (CI). By introducing the CI, the RNN-T model can overcome the homophone problem while utilizing the pronunciation information for extracting modeling units. With the proposed encoding, we changed the one-to-many mapping problem to one-to-one mapping in the inference stage so that the model outputs can be converted into the final recognition result through a simple lookup operation.
We conducted experiments on Aishell and MagicData datasets, and the experimental results showed the effectiveness of the proposed method.

\section{ACKNOWLEDGMENTS}
\label{sec:ack}
This work is supported by JSPS KAKENHI No.21K17776.

\bibliographystyle{IEEEbib}
\bibliography{mybib4asr}

\end{document}